\documentclass[letterpaper]{article} 
\usepackage[submission]{main_style}  
\usepackage{times}  
\usepackage{helvet}  
\usepackage{courier}  
\usepackage[hyphens]{url}  
\usepackage{graphicx} 
\usepackage{natbib}  
\usepackage{caption} 
\usepackage{algorithm}
\usepackage{algorithmic}
\usepackage{multirow} 
\usepackage{makecell}
\usepackage{booktabs}
\usepackage{amssymb}
\definecolor{cvprblue}{rgb}{0.21,0.49,0.74}
\usepackage[pagebackref,breaklinks,colorlinks,allcolors=cvprblue]{hyperref}
\usepackage{bookmark}

\usepackage{newfloat}
\usepackage{listings}
\DeclareCaptionStyle{ruled}{labelfont=normalfont,labelsep=colon,strut=off} 
\floatstyle{ruled}
\newfloat{listing}{tb}{lst}{}
\floatname{listing}{Listing}
\title{Turbo-VAED: Fast and Stable Transfer of Video-VAEs to Mobile Devices}
\author{
    \textbf{Ya Zou}\equalcontrib,
    \textbf{Jingfeng Yao}\equalcontrib,
    \textbf{Siyuan Yu}, 
    \textbf{Shuai Zhang}, 
    \textbf{Wenyu Liu}, 
    \textbf{Xinggang Wang}\thanks{Corresponding author: xgwang@hust.edu.cn}
}
\affiliations{
    Huazhong University of Science and Technology
}

\usepackage{bibentry}

\newcommand{\modelname}{Turbo-VAED}

\begin{document}
\makeatletter

\g@addto@macro\@maketitle{
\begin{figure}[H]
\setlength{\linewidth}{\textwidth}
\setlength{\hsize}{\textwidth}
\centering
\vspace{-11.6mm}
\includegraphics[width=\textwidth]{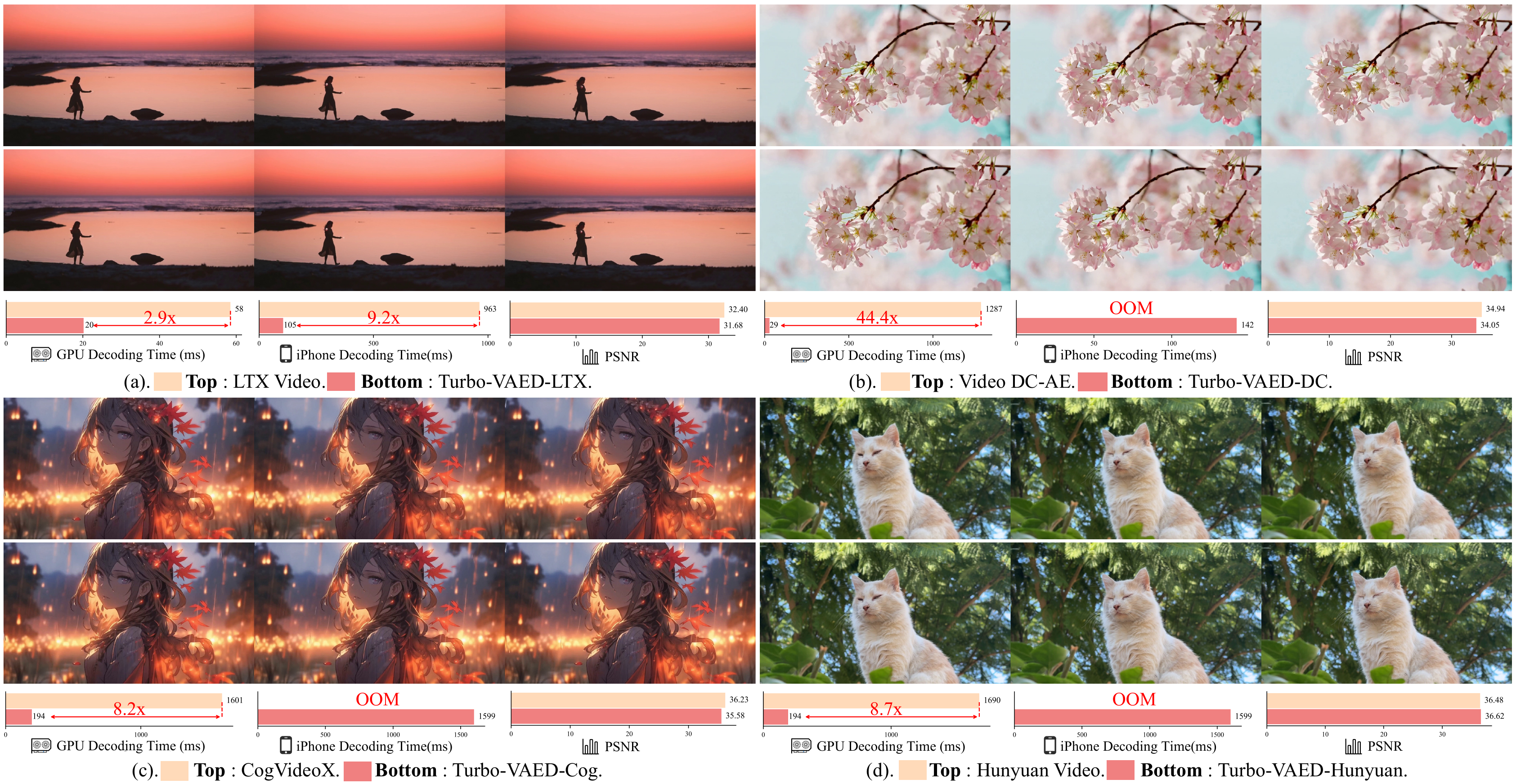} 
\caption{\textbf{Video Reconstruction Results.} We compare four widely used video VAEs and their ``turbo-charged" variants. The upper section displays reconstructed videos (Top: base model; Bottom: Turbo-VAED version). 
The turbo-charged decoders significantly reduce decoding latency across both GPU and iPhone platforms while maintaining reconstruction quality.
}   
\label{fig:main}
\end{figure}
}

\maketitle

\begin{abstract}

There is a growing demand for deploying large generative AI models on mobile devices. For recent popular video generative models, however, the Variational AutoEncoder (VAE) represents one of the major computational bottlenecks. Both large parameter sizes and mismatched kernels cause out-of-memory errors or extremely slow inference on mobile devices. 
To address this, we propose a low-cost solution that efficiently transfers widely used video VAEs to mobile devices. 
(1) We analyze redundancy in existing VAE architectures and get empirical design insights. By integrating 3D depthwise separable convolutions into our model, we significantly reduce the number of parameters. 
(2) We observe that the upsampling techniques in mainstream video VAEs are poorly suited to mobile hardware and form the main bottleneck. In response, we propose a decoupled 3D pixel shuffle scheme that slashes end-to-end delay. 
Building upon these, we develop a universal mobile-oriented VAE decoder, \textbf{\modelname{}}.
(3) We propose an efficient VAE decoder training method. 
Since only the decoder is used during deployment, we distill it to \modelname{} instead of re-training the full VAE, enabling fast mobile adaptation with minimal performance loss.
To our knowledge, our method enables \textit{real-time 720p video VAE decoding on mobile devices for the first time}. 
This approach is widely applicable to most video VAEs. 
When integrated into four representative models, with training cost as low as \$95, it accelerates original VAEs by up to 84.5$\times$ at 720p resolution on GPUs, uses as low as 17.5\% of original parameter count, and retains 96.9\% of the original reconstruction quality.
Compared to mobile-optimized VAEs, \modelname{} achieves a 2.9$\times$ speedup in FPS and better reconstruction quality on the iPhone 16 Pro.
The code and models will soon be available at \url{https://github.com/hustvl/Turbo-VAED}.

\end{abstract}

\section{Introduction}

Driven by the growing demand for deploying large generative AI models on mobile devices~\cite{smolvlm, minicpm, gemini}, adapting video generation models for mobile platforms has attracted considerable attention~\cite{snapgen, mobile-video-diffusion}.
As a key component in latent diffusion models~\cite{ldm}, VAEs~\cite{vae} compress visual signals into latent spaces. However, most current video VAEs are incompatible with mobile devices.

The pursuit of better visual compression capability has driven VAEs to scale up. 
For instance, LTX-VAE~\cite{ltx} and Video DC-AE~\cite{opensora2} reach over four times the size of SVD-VAE~\cite{svd}.
Large model sizes always cause out-of-memory (OOM) errors on mobile devices. Additionally, incompatible operators result in unacceptably slow inference. 
The 3D pixel shuffle module is widely adopted in video VAEs~\cite{ltx,h3ae} for upsampling. 
However, it suffers from poor mobile compatibility, exhibiting a standalone latency that is 11$\times$ greater than that of our mobile-optimized operator.
Consequently, lightweight models incorporating mobile-optimized operations are required to enable real-time inference. 

While training a lightweight VAE from scratch is a potential solution, it demands substantial computational resources. Moreover, compact models learn latent distributions that are markedly inferior to those of larger models. 
To address this, the decoder-only distillation method~\cite{snapgen} provides a viable direction through initial research, with room for further in-depth analysis.

In this paper, we propose \textbf{\textit{\modelname{}}}, a family of lightweight VAE decoders optimized for mobile deployment.  
Its architecture effectively reduces model redundancy and parameter count,
while our mobile-friendly upsampling strategy substantially reduces on-device inference latency. 
Our comprehensive experiments and analysis of the decoder-only distillation method, while methodologically straightforward, yield key empirical insights enabling efficient and generalizable transfer of video VAEs to mobile devices.
Specifically, we conduct the following work:

\paragraph{Mobile Model Design (Sec.~\ref{model_section}~\ref{upsample_section})}
We design a universal mobile VAE decoder incorporating the following key insights.
\textit{(1) Parameter-efficient Decoder}. Through experiments and analysis, we identify significant parameter redundancy in low-resolution layers of the VAE decoder. Integrating 3D depthwise separable convolutions into these layers substantially reduces model parameters while maintaining reconstruction quality.
\textit{(2) Mobile-friendly 3D Upsampling Strategy}. The two widely used 3D upsampling techniques are 3D pixel shuffle (high-quality but slow) and 3D interpolation (low-quality and unsupported on mobile devices). To accelerate execution speed while retaining the reconstruction quality as high as possible, we modify the 3D pixel shuffle by decoupling its spatial and temporal components.

\paragraph{Training Method (Sec.~\ref{method_section})}
Our training pipeline involves two main designs. \textit{(1) Decoder-only Distillation}. Our approach involves freezing the pre-trained VAE encoder and training a tiny decoder, preserving the high-quality latent representations unchanged. We adopt this strategy because text-to-video generation relies exclusively on the decoder to transform latents into videos. Furthermore, during diffusion model training, the encoder runs only once to convert the dataset into stored latents, while the decoder is executed repeatedly.
\textit{(2) High Data Efficiency and Negligible Cost via Feature Alignment}. We distill knowledge from the original decoder into the lightweight decoder by aligning its intermediate features. Our experiments show that training with this technique remains feasible even on limited datasets, requiring a cost as low as \$95.

\paragraph{\modelname{} Family (Sec.~\ref{teacher_section})}
To validate the broad generalizability of our model design and training method, we adopt Hunyuan-VAE~\cite{hunyuanvideo}, CogVideoX-VAE~\cite{cogvideox}, Video DC-AE~\cite{opensora2}, and LTX-VAE~\cite{ltx} as teacher models. Their corresponding student models are named \modelname-Hunyuan, \modelname-Cog, \modelname-DC, and \modelname-LTX, respectively.
\paragraph{Evaluation (Sec.~\ref{evaluation_section})}
We extensively evaluate \modelname{}. 
By reducing the parameter count to as low as 17.5\% of the original model, the \modelname{} family achieves up to a 44.4$\times$ speedup at 512px resolution and 84.5$\times$ speedup at 720p resolution on the GPU. 
While achieving acceleration, they preserve up to 96.9\% reconstruction performance and up to 97.3\% generation performance.
The lightweight design also enables the mobile deployment of previously incompatible large-scale models.
Compared to mobile-optimized video VAEs like H3AE~\cite{h3ae}, \modelname-DC achieves a 2.9$\times$ speedup in FPS and better reconstruction quality under the same compression ratio on the iPhone 16 Pro. 
Notably, \modelname-DC and \modelname-LTX enable the \textbf{\textit{first}} successful decoding of 720p videos on the iPhone at up to 38.1 FPS.
 \\

Our contributions are summarized as follows:
\begin{itemize}
\item We propose a universal mobile-oriented video VAE architecture design, featuring a parameter-efficient decoder and a mobile-friendly 3D upsampling strategy.
\item We present an efficient distillation method for transferring video VAEs to mobile devices, with total training cost as low as \$95.
\item We evaluate our method on four state-of-the-art video VAEs. The \modelname{} family reduces the parameter count to as low as 17.5\% of the original model, achieving up to 84.5$\times$ faster inference at 720p resolution on GPUs and maintaining up to 96.9\% reconstruction performance. 
Our method enables the first real-time 720p video VAE decoding on the iPhone 16 Pro.
\end{itemize}

\section{Related Work}

\subsection{Mobile Deployment of Large Models}

The demand for deploying large models on mobile devices, such as large language models (LLMs) and diffusion models~\cite{ldm,dit,fasterdit}, is increasing.
For instance, LLMs~\cite{minicpm,mobilellm,gemini,smolvlm} achieve real-time on-device execution. 
\cite{snapgen,mobile-video-diffusion,On-device-Sora} explore text-to-video generation for mobile devices.
However, deploying video diffusion models on mobile platforms remains a challenge. A critical bottleneck lies in the VAE, which cause OOM errors or extremely slow inference. And retraining compact VAEs demands significant computational resources. To bridge this gap, we propose \modelname{}, a family of lightweight VAE decoders optimized for mobile deployment.

\subsection{Video Autoencoders}
Standard autoencoder~\cite{autoencoders} learns latent representations for reconstruction, while VAE introduces probabilistic modeling via latent distribution constraints. VQ-VAE~\cite{vqvae} employs codebook-based representation discretization, and VQGAN~\cite{vqgan} integrates adversarial training~\cite{gan}. 
These autoencoders~\cite{dcae,vavae} underpin modern diffusion models by compressing pixel data into latents for efficient denoising. Notably, the community has proposed numerous high-performance video VAEs.
Some models~\cite{magvitv2} learn the distribution of discrete tokens. In contrast, most video VAEs model continuous latents. Early methods like~\cite{svd} focus on spatial compression, while later works~\cite{opensora,moviegen,vitok,cvvae,videovaeplus,reducio} compress spatial and temporal dimensions for greater redundancy reduction.
Recently, some models explore efficient inference~\cite{leanvae,cosmos,h3ae}.
However, most high-quality models still fail to achieve real-time video decoding on mobile devices. 
We explore mobile-oriented model design and efficient transfer strategies, distilling these models into the \modelname{} family.

\section{Method}

In this section, we first propose our designs based on \textit{parameter-efficient decoder} and \textit{mobile-friendly 3D upsampling strategy}, which are universal for most video VAEs. Additionally, we introduce a fast distillation method and highlight its critical role during the training process.

\subsection{Preliminary}

\paragraph{Video VAEs} To enable simultaneous compression of both videos and images into a unified latent space, most VAEs impose specific constraints on the number of input video frames. Given a video $X\in \mathbb{R}^{3\times (T+1)\times H\times W}$, the VAE encodes it into a latent representation $L\in \mathbb{R}^{C\times(\frac{T}{d_{t} }+1)\times \frac{H}{d_{h} } \times \frac{W}{d_{w} } }$, where $d_t$, $d_h$, and $d_w$ denote the downsampling factors for time, height, and width respectively.

\paragraph{{3D Depthwise Separable Convolution}}
Depthwise separable convolutions reduce computational cost and model size, enabling efficient deployment on resource-constrained devices~\cite{mobilenets,mobilenetv2}.
The 3D depthwise separable convolution (3D DW Conv) is extended to 3D vision tasks and can be described as follows~\cite{3ddwconv}:
\begin{equation}
    {\hat{\mathbf{G}}}_{k, l, t, m}=\sum_{i, j, f} {\hat{\mathbf{K}}}_{i, j, f, m} \cdot {\mathbf{F}}_{k+i-1, l+j-1, t+f-1, m}
\end{equation}
\begin{equation}
    {\mathbf{G}}_{k, l, t, n}=\sum_{i, j, f, m} {\mathbf{K}}_{i, j, f, m, n} \cdot {\mathbf{\hat{\mathbf{G}}}}_{k+i-1, l+j-1, t+f-1, m}
\end{equation}

\subsection{Reducing Parameter Redundancy}
\label{model_section}

\begin{figure}[t]
\centering
\includegraphics[width=0.9\linewidth]{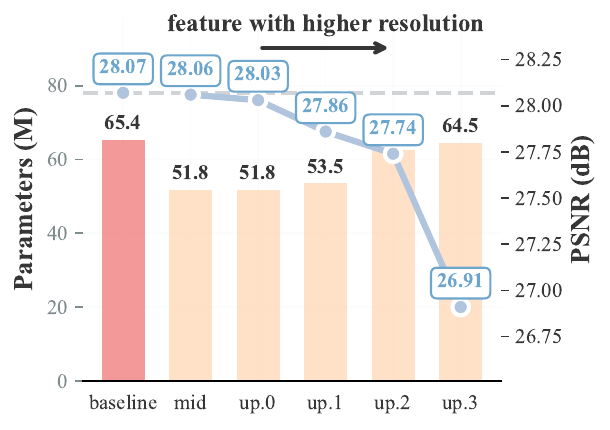}
\caption{\textbf{Decoder Redundancy Analysis.} Experimental results demonstrate that lightweight modifications at higher feature resolutions yield less substantial parameter reduction and markedly degraded reconstruction performance.}
\label{fig:redundancy}
\end{figure}

\subsubsection{Decoder Redundancy Analysis}
To improve the parameter efficiency of the decoder network, we can replace naive 3D convolutions with depthwise separable convolutions across different layers. 
We start with a lightweight decoder composed entirely of standard 3D convolutions as the baseline to distill the LTX-VAE.
We gradually apply the replacement from low-resolution, deep layers (e.g., $mid$, $up_0$) to high-resolution, top layers (e.g., $up_3$), and the results are shown in  Figure~\ref{fig:redundancy}.
The experimental results show that applying lightweight modifications from low-resolution to high-resolution layers causes a gradual increase in parameter count toward the baseline, but the reconstruction quality progressively deteriorates, as indicated by the decreasing PSNR. 
This suggests that there are many redundant parameters in the low-resolution layers, but few in the high-resolution layers.

\subsubsection{Finding 1:}In the VAE decoder, network layers processing lower-resolution features exhibit higher parameter redundancy; employing depthwise separable convolutions in these layers significantly enhances parameter efficiency.

\subsubsection{Parameter-efficient Decoder}

Our mobile decoder \modelname{} adopts a hybrid architecture, employing 3D depthwise separable convolutions in low-resolution layers and standard 3D convolutions in other layers.
We perform replacements in $mid$ and $up_0$ layers, achieving a 41.6\% reduction in parameters while maintaining virtually identical reconstruction performance (PSNR 28.05 vs. baseline 28.07).

\subsection{Accelerating 3D Upsampling}
\label{upsample_section}
\subsubsection{Mobile Upsampling Latency Analysis}

\begin{figure}[t]
\centering
 \includegraphics[width=0.9\linewidth]{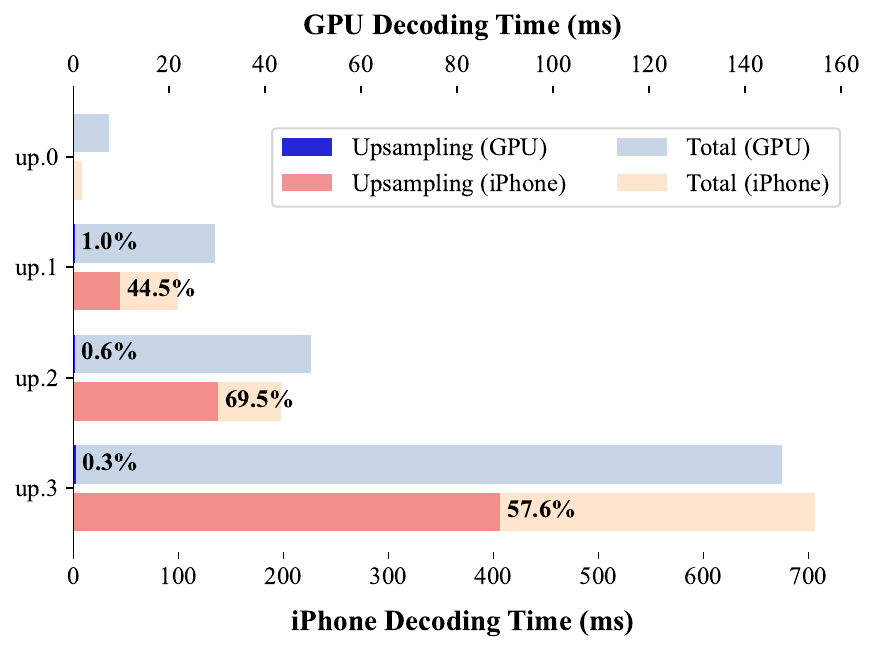}
 \caption{\textbf{Decoding Time in Different Blocks.} We conduct a thorough decoding time analysis per block. On mobile devices, the upsampling operation (3D pixel shuffle) incurs significant latency due to poor kernel compatibility, becoming the primary bottleneck in the decoding pipeline.}
 \label{fig:proportion}
\end{figure}

The 3D pixel shuffle is widely used for upsampling in video VAEs~\cite{ltx,h3ae}. 
Given its ability to achieve superior reconstruction quality (Table~\ref{tab:upsampling}), we initially incorporate it into our mobile decoder design. However, this model exhibits high inference latency on mobile devices. Therefore, we perform an in-depth decoding time analysis for each block, as shown in Figure~\ref{fig:proportion}. On GPUs, the execution time of 3D pixel shuffle accounts for a very small fraction of the decoding time per block. However, on mobile devices, it dominates the decoding time. 
This high-latency upsampling operation is the key factor that slows down the entire model’s on-device decoding speed.

\subsubsection{Finding 2:}The 3D Pixel Shuffle demonstrates low computational efficiency for upsampling on mobile devices due to poor kernel compatibility, emerging as the primary latency bottleneck during decoding.

\begin{table}[t]
\centering
\setlength{\tabcolsep}{1.25mm}
{
\begin{tabular*}{\linewidth}{lccccc}
   \toprule
   \textbf{Upsampling} & \textbf{\makecell{ Decoding \\ Time}} & \textbf{PSNR}↑ & \textbf{LPIPS}↓ & \textbf{SSIM}↑ \\
   \midrule
   3D Pixel Shuffle & 1343 ms & 28.05 & 0.1293 & 0.8431 \\
   3D Interpolate & / & 27.40 & 0.1392 & 0.8272 \\
   Ours & 446 ms & 27.86 & 0.1312 & 0.8396 \\
   \bottomrule
\end{tabular*}
}
\caption{\textbf{Upsampling Techniques.} 
We ablate different upsampling methods in the decoder architecture. Our approach achieves a balance between decoding speed and reconstruction quality. } 
\label{tab:upsampling} 
\end{table}

\subsubsection{Mobile-friendly 3D Upsampling Strategy}
Although 3D interpolation is a common alternative, it exhibits inferior reconstruction quality and lacks support in major mobile operator libraries. 
To achieve a decoder with fast inference speed, we propose a novel mobile-friendly upsampling solution.

\begin{figure}[t]
\centering
\includegraphics[width=\linewidth]{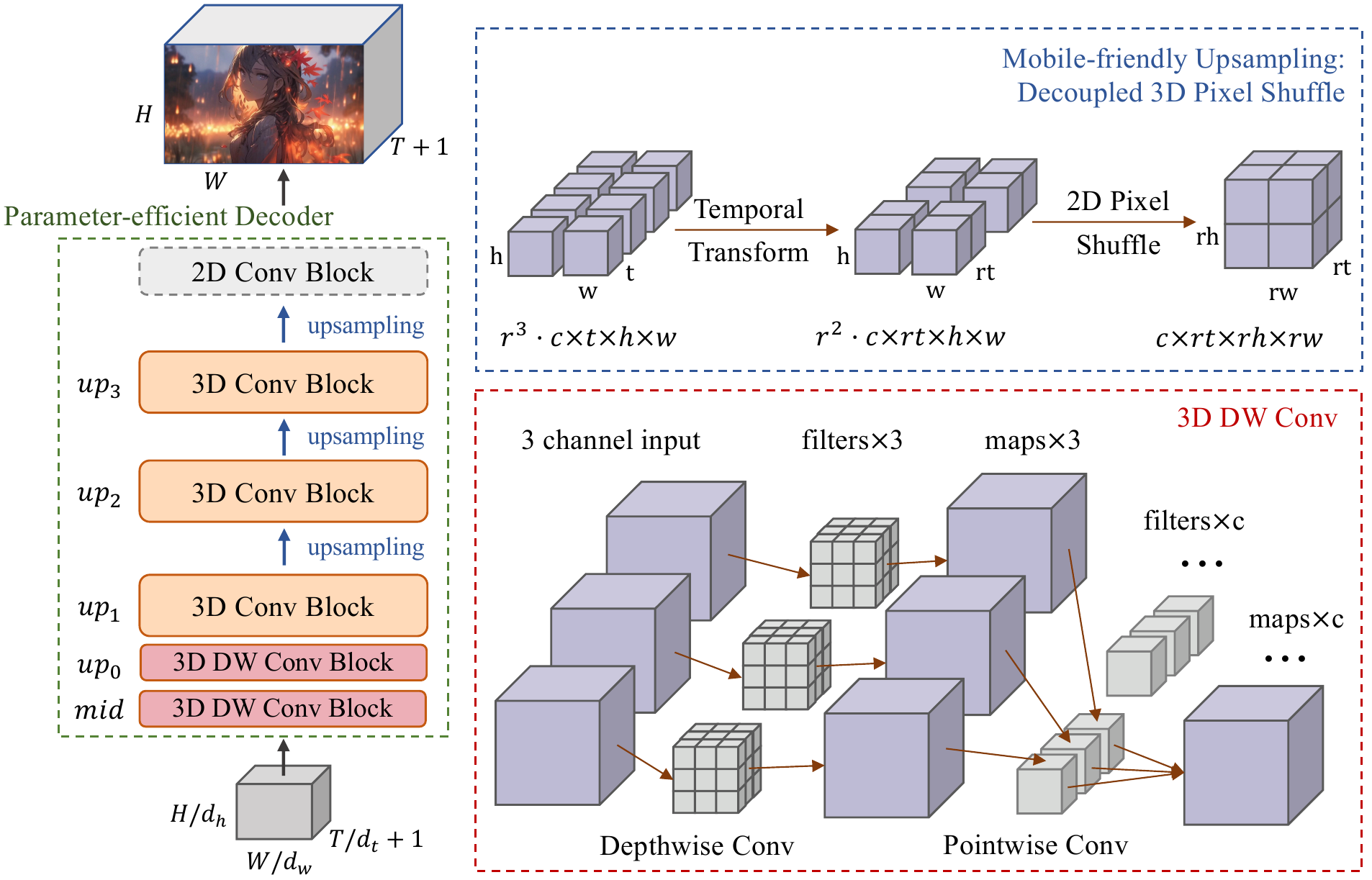}
\caption{\textbf{\modelname{} Architecture Overview.} We illustrate the mobile-oriented architecture design: a parameter-efficient decoder that incorporates a mobile-friendly 3D upsampling strategy.}
\label{fig:model}
\end{figure}

We decompose the 3D pixel shuffle into distinct temporal and spatial operations, as illustrated in the top-right of Figure~\ref{fig:model}. 
First, transform the convolution layers' output $F \in \mathbb{R}^{(r^3 \times C) \times T \times H \times W}$ by converting channels to the temporal dimension, producing an intermediate feature $\hat{F} \in \mathbb{R}^{(r^2 \times C) \times rT \times H \times W}$, where $r$ is the scaling factor. 
The spatial upsampling process involves applying 2D pixel shuffle~\cite{pixelshuffle}, which can be formulated as follows to produce the final video $Y \in \mathbb{R}^{C \times rT \times rH \times rW}$:
\begin{equation}
    Y_{c,t,h,w}=\hat{F}_{C \cdot r \cdot \bmod (w, r)+C \cdot \bmod (h, r)+c, t, \lfloor h / r\rfloor, \lfloor w / r\rfloor}
\end{equation}

Our upsampling technique results in a significantly shortened execution chain of operators after compilation, leading to faster inference speed on mobile devices.
As shown in Table~\ref{tab:upsampling}, experiments validate that \modelname{}-LTX with our upsampling technique achieves a \textit{66.8\%} speedup compared to its counterpart with 3D pixel shuffle on iPhone devices.
While our method shows slightly inferior reconstruction quality compared to 3D pixel shuffle, it outperforms 3D interpolation.
Therefore, we adopt this mobile-friendly design as the 3D upsampling strategy in \modelname{}.

\subsection{Enhancing Training Efficiency}
\label{method_section}

\subsubsection{Distillation Loss Analysis}
To obtain an efficient training method that transfers the pre-trained video VAEs to mobile devices, we employ knowledge distillation from the original decoder to \modelname{}.
Following prior knowledge distillation works~\cite{vitkd, dmae, mgd, reffakd, deit}, we design a distillation loss that aims to align the intermediate layer features of the two decoders, as defined in Equation~\ref{equ:distill}.
\begin{equation}
\label{equ:distill}
L_{distill}=\sum_{l} \frac{1}{numel\left(f_{l}^{T}\right)} \sum_{i}\left\|\sigma\left(f_{l}^{S}\right)_{i}-f_{l, i}^{T}\right\|_{1}
\end{equation}
Where $l$ denotes the number of blocks, ${numel}(\cdot)$  represents the total number of elements, $f_{l}^{T}$ and $f_{l}^{S}$ denote the features of the corresponding layers in the teacher and student decoders. And $\sigma(\cdot)$ refers to the projection network function, which maps student features to align with the teacher model's hidden dimension.

\begin{figure}[t]
\centering
\includegraphics[width=\linewidth]{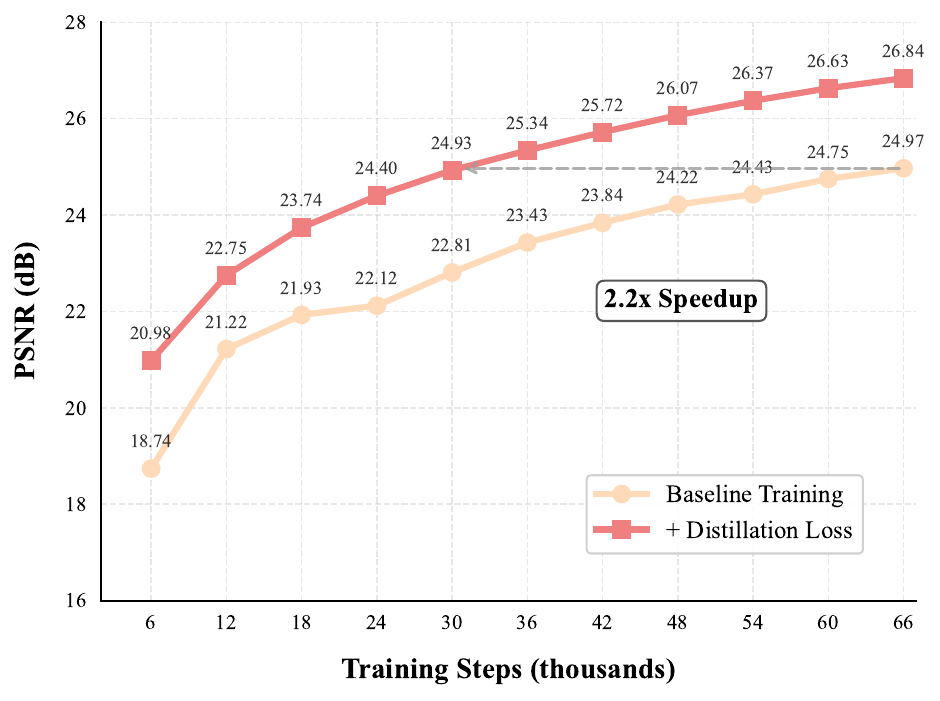}
\caption{\textbf{Distillation Loss.} We train \modelname{}-LTX on VidGen dataset at 256px resolution, ablating the additional distillation loss. The distillation loss significantly accelerates convergence while enhancing reconstruction quality.}
\label{fig:distilation}
\end{figure}

As shown in Figure~\ref{fig:distilation}, incorporating $L_{distill}$ accelerates convergence with a 2.2$\times$ speedup. 
Training \modelname{}-LTX with $L_{distill}$ yields a PSNR of 30.39 at convergence on the VidGen test set (baseline: 28.77), demonstrating superior reconstruction quality.
Furthermore, Table~\ref{tab:size} highlights that models trained on 10k and 1M video datasets using our distillation loss achieve comparable performance.

\subsubsection{Finding 3:}Feature alignment-based distillation enables data-efficient training, substantially enhancing model performance while accelerating convergence.

\begin{table}[t]
\centering
\setlength{\tabcolsep}{2.5mm}
{
\begin{tabular*}{\linewidth}{llccc}
   \toprule
   \textbf{Dataset} & \textbf{Samples} & \textbf{PSNR}↑ & \textbf{LPIPS}↓ & \textbf{SSIM}↑ \\
   \midrule
   Subset & 10,000 & 29.21 & 0.0943 & 0.8709 \\
   Full & 1,000,000 & 29.23 & 0.0950 & 0.8711 \\
   \bottomrule
\end{tabular*}
}
\caption{\textbf{Number of Training Samples.} We investigate training with our distillation loss across varying dataset sizes. Performance with 10K and 1M samples is comparable, demonstrating our method's low data requirements and high practical value.} 
\label{tab:size} 
\end{table}

\subsubsection{Efficient Distillation Method}
\begin{figure}[t]
\centering
\includegraphics[width=0.9\linewidth]{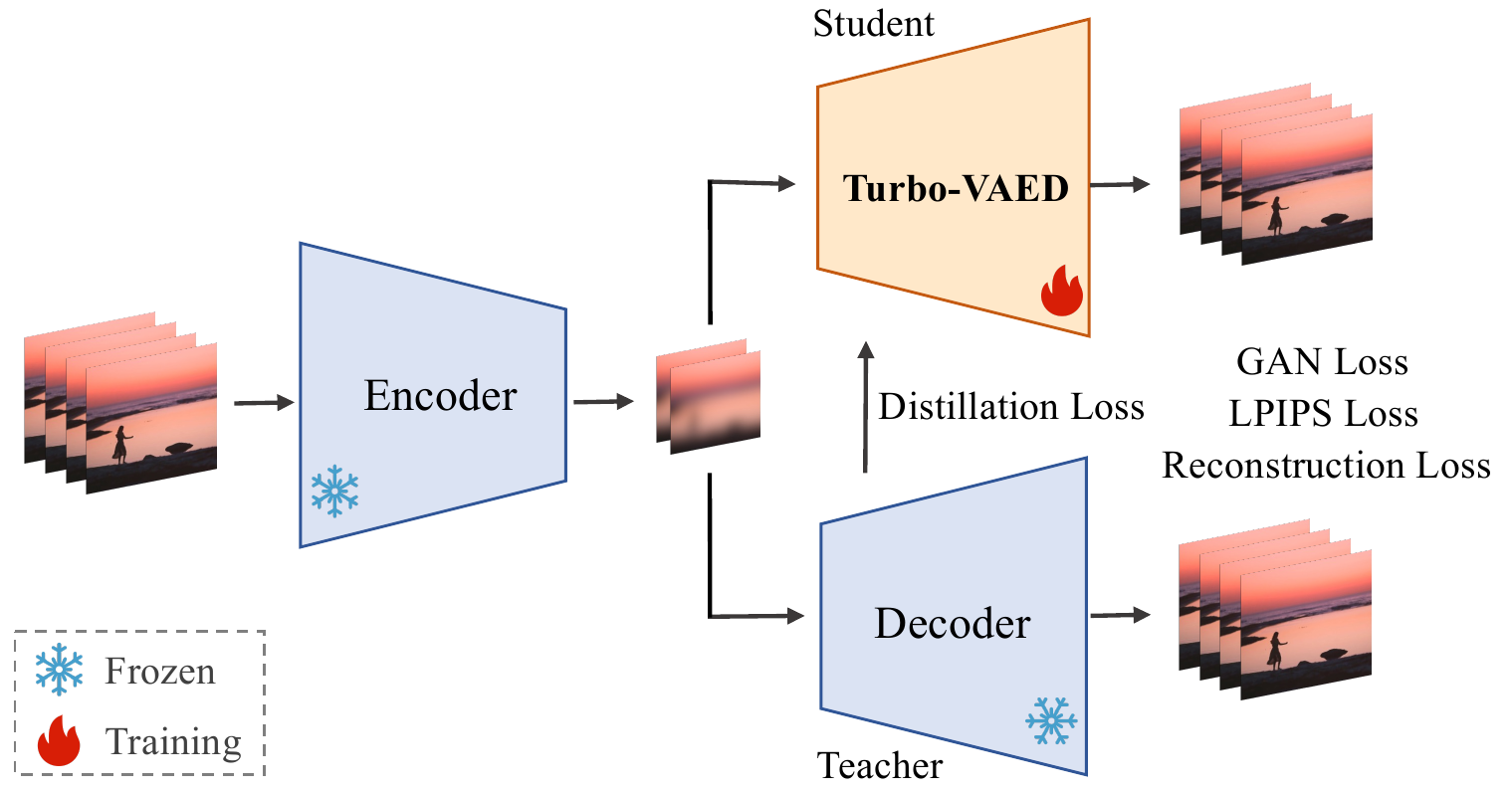}
\caption{\textbf{Training Pipeline.} The pre-trained VAE remains frozen as we distill the lightweight decoder \modelname{} by aligning the intermediate features.}
\label{fig:pipeline}
\end{figure}

As illustrated in Figure~\ref{fig:pipeline}, we freeze the encoder and distill knowledge from the original decoder to \modelname{} by aligning intermediate layer features between them.
In addition to standard reconstruction loss $L_1$ and KL loss $L_{kl}$, we incorporate the perceptual loss $L_{lpips}$, the adversarial GAN loss $L_{adv}$, and our designed distillation loss $L_{distill}$. The complete loss function is shown in Equation~\ref{equ:loss}. Following the training strategy of \cite{opensora2}, we employ a two-stage procedure: $L_{adv}$ is excluded during the initial stage and introduced only after the model reaches near-convergence in the previous stage.
\begin{equation}
\label{equ:loss}
L=L_1+{\alpha}_1 L_{lpips}+{\alpha}_2 L_{distill}+{\alpha}_3 L_{kl}+{\alpha}_4 L_{adv}
\end{equation}

\begin{table*}[t]
\centering
 \setlength{\tabcolsep}{1.55mm}
{
 \begin{tabular*}{\textwidth}{lclccccccc}
  \toprule
  \multirow{2}*{\textbf{Model}} & \multirow{2}*{\textbf{\makecell{Decoder \\ Param(M)}}} & \multirow{2}*{$\mathbf{(d_t, d_h, d_w)}$}  & \multicolumn{2}{c}{\textbf{FPS@512$\times$512}\,$\uparrow$} &  \multicolumn{4}{c}{\textbf{UCF-101@256$\times$256}} & \textbf{OpenVid} \\
  \cmidrule(lr){4-5}\cmidrule(lr){6-9}\cmidrule(lr){10-10}
  & & & \textbf{GPU} & \textbf{iPhone} &  \textbf{PSNR}↑ & \textbf{SSIM}↑ & \textbf{LPIPS}↓  & \textbf{rFVD}↓ & \textbf{FVD}↓ \\
  \midrule
HunyuanVideo     & 146.1  & ~(4,~8,~8)     & 10.1   & {\textit{OOM}}   & 36.48  & 0.9663  & 0.0126 & 1.52   &  305.38 \\
\modelname{}-Hunyuan  & \textbf{40.7}      & ~(4,~8,~8)     & 87.5  &    \textbf{10.6} & 36.62   & 0.9674  & 0.0154  &    2.43  &    306.74     \\
\midrule
CogVideoX      & 123.4 & ~(4,~8,~8)     & 10.6  & {\textit{OOM}}    & 36.23   & 0.9591 & 0.0197  & 4.73  &    254.67     \\
\modelname{}-Cog   & \textbf{40.7}    & ~(4,~8,~8)     & 87.5      &     \textbf{10.6}  & 35.58  & 0.9606  &  0.0181  &  3.09  &  278.78  \\
\midrule
Video~DC-AE       & 239.0  & ~(4,~32,~32)  & 12.4   & {\textit{OOM}}  & 34.94 & 0.9594 & 0.0196  & 4.74 & 216.07  \\
\modelname{}-DC    & \textbf{45.8}    & ~(4,~32,~32)  & 552.5   &  \textbf{112.7}  & 34.05  & 0.9475 & 0.0266  &  6.44  &    219.53     \\
\midrule
LTX Video    & 238.8  & ~(8,~32,~32)  & 290.6  & 17.7  & 32.40 & 0.9192  & 0.0394  & 25.86  &  178.82   \\
\modelname{}-LTX   & \textbf{41.9}      & ~(8,~32,~32)  & 841.6  & \textbf{161.8}  & 31.68  & 0.9209 & 0.0419 &  25.01 &  178.69 \\
  \bottomrule
\end{tabular*}
 }
\caption{\textbf{Comparison with Recent Video VAEs.} We evaluate our proposed architecture (Sec.~\ref{model_section}~\ref{upsample_section}) and training strategy (Sec.~\ref{method_section}) on four state-of-the-art video VAEs. Our method significantly reduces computational costs, with parameter counts reduced by up to 82.5\%, effectively addressing the \textit{OOM} issue, while preserving reconstruction and generation performance.}
\label{tab:main} 
\end{table*}

\begin{table*}[t]
\centering
\setlength{\tabcolsep}{2.25mm}
\begin{tabular*}{\textwidth}{lcccccccc}
  \toprule
  \multirow{2}{*}{\textbf{Model}} & \multirow{2}{*}{\textbf{Compression}} &
  \multicolumn{2}{c}{\textbf{FPS@512$\times$512}\,$\uparrow$} &
  \multicolumn{2}{c}{\textbf{FPS@720$\times$1280}\,$\uparrow$} &
  \multicolumn{3}{c}{\textbf{DAVIS@512$\times$512}} \\
  \cmidrule(lr){3-4}\cmidrule(lr){5-6}\cmidrule(lr){7-9}
  &  & GPU & iPhone & GPU & iPhone &
    \textbf{rFVD}\,$\downarrow$ & \textbf{PSNR}\,$\uparrow$ & \textbf{SSIM}\,$\uparrow$ \\
  \midrule
  SnapGen-V~\cite{snapgen}        & 1:192   & --     & 31.5 &
                   -- & -- & --     & --    & --      \\
  H3AE~\cite{h3ae}             & 1:96     & 195.4  & 38.1 &
                   -- & -- & 122.82 & \textbf{30.23} & 0.8412 \\
  \midrule
  \modelname{}-LTX & 1:192    &
    \textbf{841.6} & \textbf{161.8} &
 \textbf{255.6} & \textbf{38.1} & 125.28 & 27.86 & 0.7905 \\
  \modelname{}-DC  & 1:96     &
     552.5 & 112.7 &
    167.0 & 25.3 & \textbf{49.91} & 30.08 & \textbf{0.8492} \\
  \bottomrule
\end{tabular*}
\caption{\textbf{Comparison with Mobile‑optimized VAEs.} Our models achieve significantly faster inference than prior mobile-optimized models while delivering competitive reconstruction quality.}
\label{tab:iphone}
\end{table*}

\section{Experiments}

\subsection{\modelname{} Family}
\label{teacher_section}
We employ SOTA video VAEs from ~\cite{hunyuanvideo, cogvideox, ltx, opensora2} as teacher models for distillation.
Hunyuan-VAE realizes near-lossless video fidelity.
CogVideoX-VAE effectively minimizes artifacts in complex dynamic scenarios. 
Video DC-AE extends the~\cite{dcae} framework for high-ratio video compression, achieving high-quality reconstruction.
LTX-VAE achieves a high compression ratio of 1:192, preserving the ability to generate fine details.
However, these models encounter issues during mobile deployment due to their high parameters and mismatched kernels. 
So we separately distill decoders for each model to improve inference speed while striving to maintain the original high quality.

\subsection{Implementation Details}
We train our \modelname{} on a subset of the VidGen~\cite{vidgen} video dataset, consisting of 10k videos, which are preprocessed into 17-frame sequences at 256$\times$256 resolution.
We adopt the architecture from LTX-VAE as our initial decoder framework and refine it using the design techniques described in Section~\ref{model_section} and~\ref{upsample_section}.
Empirically, we set ${\alpha}_1=1.0$, ${\alpha}_2=1.0$, ${\alpha}_3=1\times {10}^{-7}$, and ${\alpha}_4=0.05$.
The training is conducted on NVIDIA V100 GPUs, totaling about 300 GPU-hours, and gradient accumulation is implemented with an effective batch size of 32.
We use AdamW optimizer with a learning rate of 2e-4 and $\beta$ set to [0.9, 0.95].

\begin{table}[t]
\centering
\setlength{\tabcolsep}{2.4mm}
{
\begin{tabular*}{\linewidth}{ccccc}
   \toprule
   \textbf{\makecell{Decoder \\ Param(M)}} & \textbf{\makecell{Kernel \\ Size}} & \textbf{PSNR}↑ & \textbf{LPIPS}↓ & \textbf{SSIM}↑ \\
   \midrule
   51.80 & 3 & 27.99 & 0.1310 & 0.8425 \\
   51.90 & 5 & \textbf{28.09} & \textbf{0.1285} & 0.8430 \\
   52.13 & 7 & 28.07 & 0.1307 & \textbf{0.8438} \\
   \bottomrule
\end{tabular*}
}
\caption{\textbf{Ablation on 3D Convolution Kernel Size.} The $5 \times 5 \times 5$ kernel size performs best.} 
\label{tab:kernel_size} 
\end{table}

\subsection{Evaluation}
\label{evaluation_section}
Following~\cite{seaweed,h3ae}, we benchmark reconstruction quality on the UCF-101~\cite{ucf101} testval and DAVIS-2017~\cite{davis2017} test datasets, reporting PSNR, LPIPS, SSIM, and reconstruction-FVD (rFVD) as evaluation metrics.
We use the FVD metric to assess text-to-video generation performance on the OpenVid~\cite{openvid} dataset at 360$\times$640 resolution.
We report decoding latency on both the NVIDIA A100 GPU and iPhone 16 Pro at 512px and 720p. 
All video datasets are used with 17 frames in standard settings and 16 frames for \modelname{}-DC during training and testing.

As shown in Table~\ref{tab:main}, the \modelname{} family retains quality with minimal degradation, and accelerates inference speed.
\modelname{}-Hunyuan achieves 8.7$\times$ speedups over Hunyuan-VAE for 512px video inference on GPUs, with slightly higher PSNR and SSIM than the original and minor trade-offs in LPIPS and FVD, demonstrating competitive reconstruction and generation performance.
Similarly, \modelname{}-Cog delivers 8.2$\times$ speedups at 512px compared to CogVideoX-VAE on GPUs while retaining comparable quality (8.1\% lower LPIPS, 34.7\% lower rFVD, 9.5\% higher FVD).
Both models enable mobile deployment at 512px resolution without OOM errors.

\modelname{}-DC delivers 44.4$\times$ and 84.5$\times$ speedups over Video DC-AE for 512px and 720p video inference on GPUs, using just 19.2\% of its parameters.
At the same 1:96 compression ratio, \modelname{}-DC achieves a 2.9$\times$ speedup in FPS over H3AE and demonstrates better reconstruction performance with a 59.4\% reduction in rFVD (Table~\ref{tab:iphone}).
While reducing parameters to 17.5\%, \modelname{}-LTX delivers a 9.2× speedup over LTX-VAE at 512px resolution on mobile devices, achieving comparable quality with slightly worse LPIPS but improved rFVD and FVD. 
For the first time, \modelname{}-DC and \modelname{}-LTX extend the capability of 720p video decoding to mobile devices, with \modelname{}-LTX achieving 38.1 FPS for this task.

\begin{figure*}[t]
\centering
 \includegraphics[width=\textwidth]{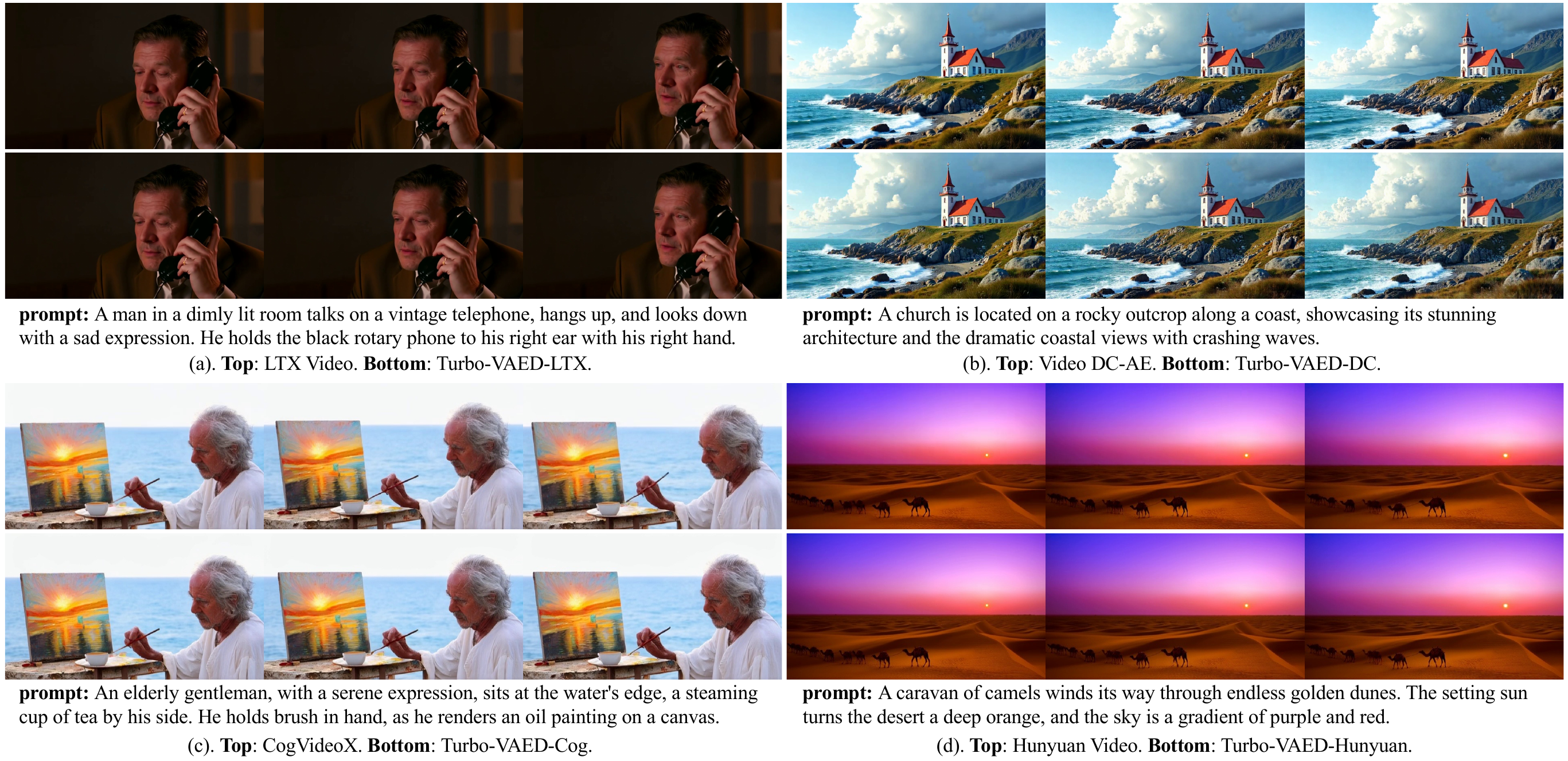}
 \caption{\textbf{Text to Video Generation Results.} The top and bottom rows display the generated videos, with latents produced by the original diffusion models and decoded via the original VAEs and their \modelname{} variants. These results with minimal visual differences demonstrate that \modelname{} preserves generation quality effectively.}
 \label{fig:generation}
\end{figure*}

\subsection{Ablations}

\paragraph{Ablation on 3D Convolution Kernel Size}

We perform ablation studies on 3D depthwise separable convolutions with different kernels.
Larger kernels enhance model performance by expanding receptive fields, while depthwise separable convolutions reduce computational costs~\cite{convnet}.
As shown in Table~\ref{tab:kernel_size}, kernel sizes of $5\times5\times5$ and $7\times7\times7$ outperform the baseline, with the former achieving the best PSNR and LPIPS.
Large kernels introduce a limited parameter increase and impose under 10~ms additional decoding latency on mobile devices.
We adopt $5\times5\times5$  kernels for 3D depthwise separable convolutions in \modelname{}.

\begin{table}[t]
\centering
\setlength{\tabcolsep}{3.1mm}
{
\begin{tabular*}{\linewidth}{lccc}
   \toprule
   \textbf{Alignment Block} & \textbf{PSNR}↑ & \textbf{LPIPS}↓ & \textbf{SSIM}↑ \\
   \midrule
   $mid$ & 26.30 & 0.1563 & 0.7972 \\
   $up_0$ & 26.46 & 0.1514 & 0.8032 \\
   $up_1$ & 26.42 & 0.1512 & 0.7992 \\
   $up_2$ & 24.82 & 0.1837  &  0.7455 \\
   \midrule
   $up_0$ \& $up_1$ & \underline{26.83} & \underline{0.1441} & \underline{0.8124} \\
   $mid$ \& $up_0$ \& $up_1$ & \textbf{26.91} & \textbf{0.1391} & \textbf{0.8155} \\
   \bottomrule
\end{tabular*}
}
\caption{\textbf{Ablation on Feature Alignment Location.} Aligning multiple layers yields better reconstruction quality.} 
\label{tab:location} 
\end{table}

\paragraph{Ablation on Feature Alignment Location}
Aligning features on different decoder blocks impacts reconstruction quality. 
As shown in Table~\ref{tab:location}, aligning low-resolution features outperforms high-resolution counterparts, achieving a 17.7\% improvement in LPIPS. Moreover, aligning multiple layers yields better results than any single-layer alignment, with an 8\% LPIPS reduction compared to the best single-layer baseline. Empirically, these findings hold across all models in our experiments, leading us to adopt the multi-layer alignment strategy in all studies.

\paragraph{Ablation on Feature Projection Head}

We analyze the impact of different projection networks for feature alignment, as shown in Table~\ref{tab:projection}. 
Feature alignment distillation employs a small projection head to project student features to match the teacher’s hidden dimension while providing extra flexibility~\cite{dmae}.
Observation indicates that a two-layer linear network built with 1$\times$1$\times$1 convolutions outperforms other configurations.

\begin{table}[t]
\centering
\setlength{\tabcolsep}{2.8mm}
{
\begin{tabular*}{\linewidth}{lccc}
   \toprule
   \textbf{Projection Head} & \textbf{PSNR}↑ & \textbf{LPIPS}↓ & \textbf{SSIM}↑ \\
   \midrule
   Linear & 26.88 & 0.1470 & 0.8148 \\
   1-layer MLP & 26.81 & 0.1424 & 0.8119 \\
   2-layer MLP & 26.80 & 0.1445 & 0.8120 \\
   3D Pointwise Conv & \textbf{26.91} & \textbf{0.1391} & \textbf{0.8155} \\
   \bottomrule
\end{tabular*}
}
\caption{\textbf{Ablation on Feature Projection Head.} The two-layer 3D pointwise convolution network is the optimal choice.} 
\label{tab:projection} 
\end{table}

\section{Conclusion}
This paper focuses on VAEs as deployment bottlenecks for video generative models on mobile devices.
To address this problem, we propose a universal mobile-oriented video VAE decoder design, featuring (1) a parameter-efficient architecture based on 3D depthwise separable convolutions and (2) a decoupled 3D pixel shuffle upsampling strategy. 
We present a data-efficient training method enabling fast and stable transfer of video VAEs to mobile devices with negligible training cost.
The solution is widely applicable to most video VAEs. 
It accelerates original VAEs by up to 84.5$\times$ at 720p resolution on GPUs, using as low as 17.5\% of the original parameter count while preserving 96.9\% of the original reconstruction quality. 
To our knowledge, \modelname{} achieves the first real-time 720p video VAE decoding on mobile devices.
Our work aims to facilitate future research on the mobile deployment of large video generative models.

\bibliography{main}

\end{document}